\documentclass[letterpaper, 10 pt, conference]{ieeeconf}  

\IEEEoverridecommandlockouts                              

\usepackage{graphics} 
\usepackage{epsfig} 
\usepackage{mathptmx} 
\usepackage{amsmath} 
\usepackage{amssymb}  
\usepackage{lipsum}
\usepackage{bm}
\usepackage{xcolor}

\usepackage[style=ieee]{biblatex}

\DeclareSourcemap{
  \maps{
    \map{
      \pertype{article}
      \step[fieldset=language, null]
      \step[fieldset=url, null]
      \step[fieldset=doi, null]
      \step[fieldset=issn, null]
      \step[fieldset=isbn, null]
      \step[fieldset=note, null]
      \step[fieldset=editor, null]
      \step[fieldset=urldate, null]
      \step[fieldset=file, null]
    }
  }
}
\DeclareSourcemap{
  \maps{
    \map{
      \pertype{inproceedings}
      \step[fieldset=language, null]
      \step[fieldset=url, null]
      \step[fieldset=doi, null]
      \step[fieldset=issn, null]
      \step[fieldset=isbn, null]
      \step[fieldset=note, null]
      \step[fieldset=editor, null]
      \step[fieldset=urldate, null]
      \step[fieldset=file, null]
    }
  }
}
\DeclareSourcemap{
  \maps{
    \map{
      \pertype{incollection}
      \step[fieldset=language, null]
      \step[fieldset=url, null]
      \step[fieldset=doi, null]
      \step[fieldset=issn, null]
      \step[fieldset=isbn, null]
      \step[fieldset=note, null]
      \step[fieldset=editor, null]
      \step[fieldset=urldate, null]
      \step[fieldset=file, null]
    }
  }
}

\title{\LARGE \bf
Conjugate momentum based thruster force estimate in dynamic multimodal robot
}

\author{Shreyansh Pitroda$^{1}$, Eric Sihite$^{2}$, Taoran Liu$^{1}$, Kaushik Venkatesh Krishnamurthy$^{1}$, \\ Chenghao Wang$^{1}$, Adarsh Salagame$^{1}$, Reza Nemovi$^{2}$, Alireza Ramezani$^{1*}$, and Morteza Gharib$^{2}$
\thanks{$^{1}$ The authors are with SiliconeSynapse Labs, the Department of Electrical Engineering, Northeastern University, USA.}%
\thanks{$^{2}$ The authors are with the Department of Aerospace Engineering, California Institute of Technology, USA.}%
\thanks{$^{*}$ The corresponding author. Email: a.ramezani@northeastern.edu}%
}

\begin{document}

\maketitle
\thispagestyle{empty}
\pagestyle{empty}


\begin{abstract}
In a multi-modal system which combines thruster and legged locomotion such our state-of-the-art Harpy platform to perform dynamic locomotion. Therefore, it is very important to have a proper estimate of Thruster force. Harpy is a bipedal robot capable of legged-aerial locomotion using its legs and thrusters attached to its main frame.  we can characterize thruster force using a thrust stand but it generally does not account for working conditions such as battery voltage. In this study, we present a momentum-based thruster force estimator. One of the key information required to estimate is terrain information. we show estimation results with and without terrain knowledge. In this work, we derive a conjugate momentum thruster force estimator and implement it on a numerical simulator that uses thruster force to perform thruster-assisted walking. 

\end{abstract}

\color{black}
\section{Introduction}

Many schemes exist for estimating external forces. A thorough overview and analysis of them is given in \cite{haddadin_robot_2017}. The most well-established method is the so-called momentum observer \cite{de_luca_sensorless_2005, de_luca_actuator_2003,cai_predefined-time_2023}. Momentum observer removes the need to estimate the state's acceleration which is generally noisy in nature. It also removes the need for inversion matrices which can lead to numerical inaccuracy. Due to these advantages, it has been widely used in many applications such as human-robot interaction \cite{garofalo_sliding_2019, noauthor_making_nodate}, estimating ground contact force in legged locomotion \cite{liu_sensorless_2024,bledt_contact_2018,noauthor_residual-based_nodate} and estimating the collision force in flying drones \cite{tomic_external_2017}. Recent work has explored the application of thruster and posture manipulation in state-of-the-art robots such as the Multi-modal mobility morphobot (M4) \cite{sihite_multi-modal_2023, sihite_efficient_2022, mandralis_minimum_2023} and LEONARDO \cite{kim_bipedal_2021,dangol_control_2021-1,liang_rough-terrain_2021, sihite_optimization-free_2021, sihite_efficient_2022,ramezani_generative_2021,dangol_feedback_2020-1}.

The M4 robot aims to enhance locomotion versatility by integrating posture control and thrust vectoring, thereby enabling walking, wheeling, flying, and loco-manipulation. Similarly, LEONARDO, a bipedal quadcopter, is capable of both quasi-static walking and flying. However, neither robot fully achieves dynamic legged locomotion and aerial mobility. The integration of these modes presents a significant challenge due to their conflicting requirements (see \cite{sihite_multi-modal_2023}).

\begin{figure}[t]
    \vspace{0.05in}
    \centering
    \includegraphics[width= 0.7\linewidth]{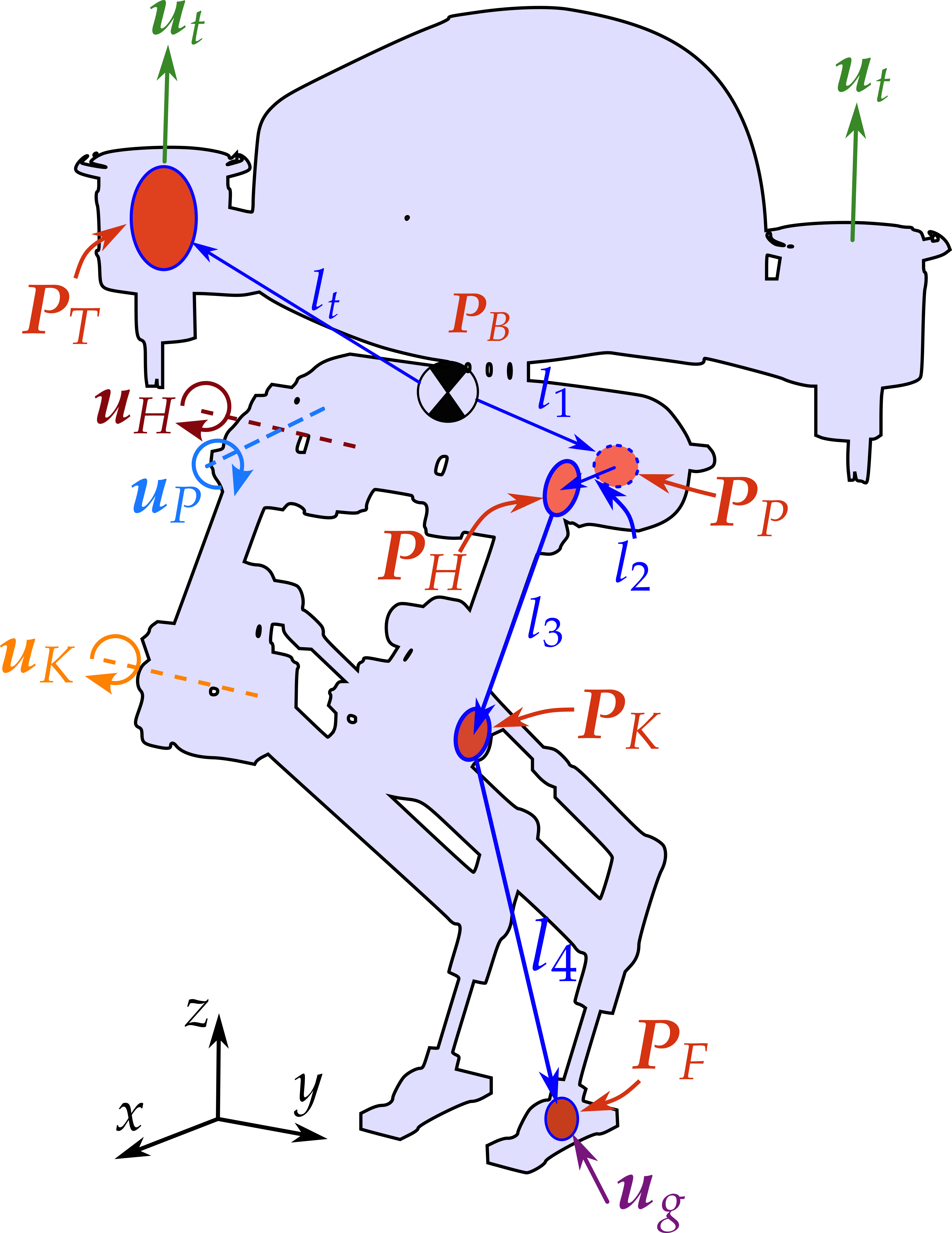}
    \caption{Illustrates the Harpy platform, a legged-aerial bipedal robot, and its kinematic chain and components.}
    \label{fig:cover-image}
    \vspace{-0.05in}
\end{figure}

To execute dynamic maneuvers, it is crucial to have an accurate estimate of the thruster force. While thruster stands can be used to characterize the generated thruster force under ideal conditions, it is challenging to account for test conditions such as battery voltage drop. Therefore, onboard thruster force estimation is essential and can be utilized by controllers such as Model Predictive Control (MPC) and Quadratic Programming (QP).

A commonly used method for estimating thruster forces in aerial manipulators involves focusing solely on the dynamics of the flying platform and treating the forces exerted by the manipulator arm as external disturbances. This approach enables the use of single-body estimation techniques to estimate thrust forces \cite{ruggiero_multilayer_2015}. However, this method is not suitable for thruster-assisted bipedal robots because the dynamics of the flying and manipulator components are coupled. Another method proposed for thrust estimation framework for the multi-link robot iRonCub \cite{mohamed_momentum-based_2022}, a multibody robot with jet propulsion attached to its back and hands, proposes an Extended Kalman Filter (EKF) based on centroidal momentum and propeller models. This framework cannot be directly implemented as their robot is in a flying state, and during contact ground reaction forces are given force sensor.

This work proposes the use of the Momentum Observer methodology to estimate the thruster force in our multimodal platform, Harpy. By implementing this approach, we aim to achieve more precise control and enhance the overall performance of the robot in dynamic environments.

In this study, we employ a detailed model of Harpy (depicted in Fig.~\ref{fig:cover-image}) using Matlab to evaluate our estimator's effectiveness. Harpy is equipped with eight custom-designed high-energy density actuators for dynamic walking, along with electric ducted fans mounted on its torso sides. Harpy's height measures 600 cm and weighs 4 Kg. It hosts a computer based on Elmo amplifiers for real-time low-level control command executions.

Harpy's design integrates advantages from both aerial and dynamic bipedal legged systems. Currently, the hardware design and assembly of Harpy have been completed~\cite{pitroda_dynamic_2023}, and our primary goal is to explore various control design strategies for this platform \cite{dangol_feedback_2020-1, dangol_control_2021,pitroda_capture_2024}.

In this work, we aim to design a conjugate momentum-based estimator to estimate thruster forces. The main contributions of this paper are: a) Formulation of Conjugate momentum-based observer for thruster forces. b) Comparison of estimate with and without knowing the terrain information.

This work is structured as follows: we present brief over view of modeling where it show dynamic model, complaint ground model and reduced-order model. Next, we present a conjugate momentum observer followed by results and conclusion.

\begin{figure}
    \vspace{0.05in}
    \centering
    \includegraphics[width=0.8\linewidth]{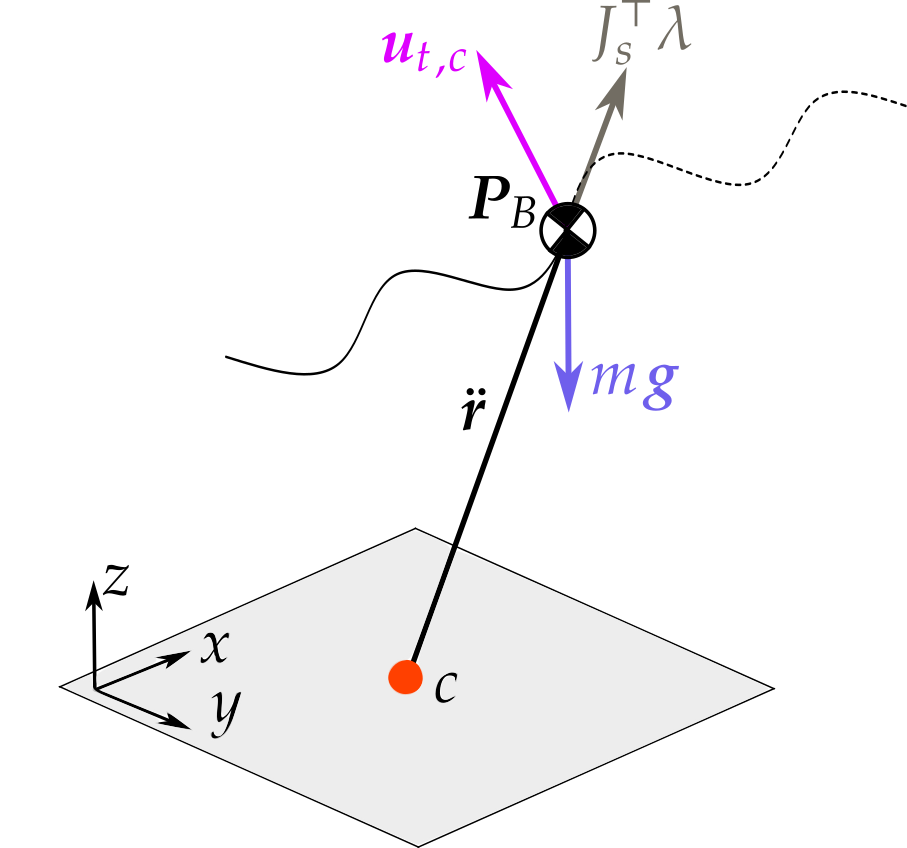}
    \caption{Illustrates Reduced-order, variable length model with a point-mass and mass-less link subjected to thruster force.}
    \label{fig:vlip_mdl}
    \vspace{-0.05in}
\end{figure}
\color{black}
\section{Dynamic modeling and control}
This section contains a brief overview of the dynamic model used in this paper for simulation, which is followed by a reduced-order model (ROM) and controller design. 

Figure~\ref{fig:cover-image} illustrates the degrees of freedom (DoF) of the robot’s leg, which includes three actuated joints: hip frontal (Pelvis $P$), hip sagittal ( hip $H$), and knee sagittal (knee $K$) joints. Combined with the robot’s body, the system has a total of 12 DoF.

The thrusters are designed to rotate about the body’s sagittal axis. However, for simplification, we assume that the thrusters can provide forces in any direction, and the thruster dynamics are ignored. The model is further simplified by assuming that the mass is concentrated at the body and the joint motors, resulting in a simpler model where the lower leg (shin and foot) is massless. The foot is considered small and modeled as a point foot, simplifying the ground force effect on the system at the cost of reduced stability due to the smaller support polygon.

The dynamic model of Harpy, used in the numerical simulation, is derived using the Euler-Lagrangian dynamic formulation. The body rotation is derived using the modified Lagrangian for dynamics in SO(3) to avoid gimbal lock or singularity present in the Tait-Bryan representation of the rotation matrix. Let $x$ be the system states, defined as follows:

\begin{equation}
    \bm x = [\bm r_{B}^\top, \bm q^\top, \phi_{K_L}, \phi_{K_R}, \bm \omega_B^{B \top}, \dot{\bm q}^\top, \dot{\phi}_{K_L}, \dot{\phi}_{K_R}]^\top,
\label{eq:states}
\end{equation}

where $\bm q = [\bm p_B^\top, \gamma_{h_L}, \gamma_{h_R}, \phi_{h_L}, \phi_{h_R}]^\top$ represents the dynamical system states other than $(R_B,\bm \omega^B_B)$. $[\gamma_{h},\phi_{h}]$ represents the hip frontal and hip sagittal joint angle in the order $[L,R]$. The knee sagittal angle $\phi_k$, which is not associated with any mass, is updated using the knee joint acceleration input $\bm u_k = [\ddot{\phi}_{k_L}, \ddot{\phi}_{k_R}]^\top$. Then, the system acceleration can be derived as follows:
\begin{equation}
\begin{gathered}
    M \bm a + \bm h = B_j\, \bm u_j + B_t\, \bm u_t + B_g\, \bm u_g
\end{gathered}
\label{eq:eom_accel}
\end{equation}

here $\bm u_j$ is the joint actuation input, $\bm u_t$ is the thruster forces and $\bm u_g$ is the GRF. each of these input are separated into the left and right components. Mapping matrix for joint is $B_j = [0_{6x6}, I_{6x6}]$. $B_t$ and $B_g$ are used to map the thruster and GRF to generalized coordinates, respectively. $\bm u_g$ and $\bm u_j$ is modelled in the further section
\begin{equation}
\begin{aligned}
    B_t = \begin{bmatrix}
        \begin{pmatrix}
        \partial \dot{\bm p}_{T_L} / \partial \bm v \\
        \partial \dot{\bm p}_{T_R} / \partial \bm v
        \end{pmatrix}^\top
        \\
        0_{2 \times 6}
    \end{bmatrix}, \quad
    B_g = \begin{bmatrix}
        \begin{pmatrix}
        \partial \dot{\bm p}_{F_L} / \partial \bm v \\
        \partial \dot{\bm p}_{F_R} / \partial \bm v
        \end{pmatrix}^\top
        \\
        0_{2 \times 6}
    \end{bmatrix}.
\end{aligned}
\label{eq:generalized_forces}
\end{equation}
\subsection{Compliant Ground Model}

For simulation, GRF is modeled using the compliant ground model with undamped rebound, while friction is modeled using the Stribeck friction model, defined as follows:
\begin{equation}
\begin{aligned}
    u_{g,z} =& -k_{g,p}\, p_{F,z} - k_{g,d}\, \dot p_{F,z} \\
    u_{g,x} =& -\left(\mu_c + (\mu_s - \mu_c)\, \mathrm{exp}\left(-\tfrac{|\dot p_{F,x}|^2}{v_s^2}\right) \right) f_z\, \mathrm{sgn}(\dot p_{F,x}) \\ 
    & - \mu_v\,\dot p_{F,x},
\end{aligned}
\label{eq:ground_model}
\end{equation}
where inertial foot position is defined 
by $p_{F,x}$ and $p_{F,z}$. Spring and damping stiffness for the ground is denoted by $k_{g,p}$ and $k_{g,d}$. $\mu_c$, $\mu_s$, and $\mu_v$ are the Coulomb, static, and viscous friction coefficients, respectively, and $v_s$ is the Stribeck velocity. $k_{g,d}$ is set to $0$ if $\dot p_{F,z} > 0$ for the undamped rebound model, and friction in the $y$ direction follows a similar derivation to $u_{g,x}$. Then, the ground force model $\bm u_g$ is defined as follows:
\begin{equation}
\begin{aligned}
    \bm u_g = [\bm u_{g_L}^\top\, H(-p_{F_L,z}),\, \bm u_{g_R}^\top\, H(-p_{F_R,z})]^\top,
\end{aligned}
\label{eq:ground_forces}
\end{equation}
where $H(x)$ denotes the Heaviside function, while $\bm u_{g_L}$ and $\bm u_{g_R}$ represent the left and right ground forces, which are formed using their respective components $u_{g,x}$, $u_{g,y}$, and $u_{g,z}$.

\subsection{Reduced-Order model}

The thruster controller was developed using the VLIP model. In this model, the center of pressure (CoP), represented as $\bm c$, is calculated as the weighted average position of the feet: $\bm c = \lambda_L\, \bm p_{F_L} + \lambda_R\, \bm p_{F_R}$, where $\lambda_i = u_{g_i,z} / (u_{g_L,z} + u_{g_R,z})$ for $i \in \{L,R\}$. In the Harpy full-dynamics model, which employs a point foot, $\bm c$ corresponds to the stance foot position during the single support (SS) phase. The VLIP model is underactuated without thrusters, but adding thrusters makes the system fully actuated, allowing for trajectory tracking. Consequently, the VLIP model is derived as follows:
\begin{equation}
\begin{gathered}
    m \ddot{\bm p}_B = m \bm g + \bm u_{t,c} + J_s^\top \bm \lambda\\ 
\end{gathered}
\label{eq:model_vlip}
\end{equation}
where $m$ represents the mass of the VLIP model, which in this case is the total mass of the system, and $\bm u_{t,c}$ denotes the thruster forces about the CoM. The constraint force $J_s^\top \bm \lambda$ is established to maintain the leg length $r$ equal to the leg conformation, utilizing the following constraint equation:
\begin{equation}
\begin{gathered}
    J_s\, (\ddot{\bm p}_B - \ddot {\bm c}) = u_r, \\  
    J_s = (\bm p_B - \bm c)^\top,
\end{gathered}
\label{eq:model_vlip_constraint}
\end{equation}
which is designed to maintain the leg length's second derivative equal to $u_r$. This constraint force also constitutes the GRF as long as the friction cone constraint is satisfied. Assuming no slip ($\ddot{\bm c} = 0$), the inputs to the system are $u_r$, which controls the body position about the vector $\bm r = \bm p_B - \bm c$ by adjusting the leg length, and the thrusters $\bm u_t$, which control the remaining degrees of freedom.

\subsection{Controller design}

The joint controller is designed to follow the desired foot positions by employing inverse kinematics to determine the target joint angles. Let $q = [\gamma_H, \phi_H, \phi_K]^{\top}$ represent the joint angles of the legs. Given the trajectory $\bm q_t$, the joint controller $\bm u_j$ can be derived using a simple PID controller. The trajectory to be tracked is developed through optimization on a $2$D version of the dynamic model shown in~\ref{fig:vlip_mdl}. However, this trajectory is unstable when applied to the full 3D system, necessitating the use of thrusters to stabilize the dynamics.
The controller uses thruster force to stabilize the roll and yaw motion of the robot according to the following law:
\begin{equation}
\begin{aligned}
    \bm u_{t,L} = [u_{yaw},0,u_{roll}],~\bm u_{t,R} = -\bm u_{t,L}
\end{aligned}
\label{eq:PD_thrust_force}
\end{equation}
Here $\bm u_{t,L}$ and $\bm u_{t,R}$ are the left and the right thruster force, respectively. $u_{yaw}$ and $u_{roll}$ are simply PD controllers to track zero roll and yaw reference angles. Combined thrust forces are formed by combining $\bm u_{t}$ in \eqref{eq:model_vlip} and \eqref{eq:PD_thrust_force} 
\begin{equation}
\begin{gathered}
    \bm u_{t} = [\bm u_{t,c}^\top,\bm u_{t,c}^\top]^\top /2 + [\bm u_{t,L}^\top,\bm u_{t,R}^\top]^\top
\end{gathered}
\end{equation}
\section{Conjugate momentum observer design}
\begin{figure}
    \vspace{0.05in}
    \centering
    \includegraphics[width=\linewidth]{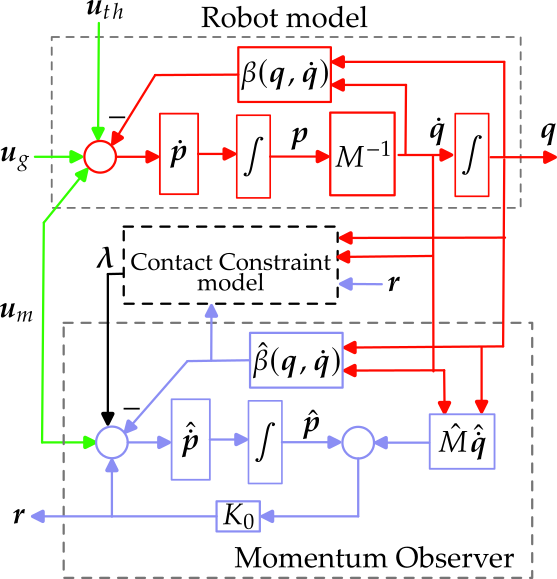}
    \caption{An overview of the estimator algorithm.}
    \label{fig:observer_flow_chart}
    \vspace{-0.05in}
\end{figure}
The approach of estimation using generalized momentum was presented in \cite{haddadin_robot_2017}, \cite{de_luca_collision_2006}, and \cite{de_luca_sensorless_2005}, driven by the aim to avoid inversion of the inertia matrix and estimation of states acceleration.n We use this approach to estimate the thruster force $\bm u_{t}$. Consider observer dynamics as, 
\begin{equation}
    \centering
    \hat M \hat{\ddot{\bm q}} + \hat { \bm h} = \hat { B}_g \, \hat{\bm \lambda} + \hat{B}_t \, \hat{\bm u_t} +  \hat {B}_j \, \hat{\bm u_j}
    \label{eq:observer-dynamics}
\end{equation}

where, $ \hat{\bm h} = \hat{G}(\hat{\bm q}) + \hat{C}\left( \hat{\bm q}, \hat{\dot{\bm q}} \right) \, \hat{\dot{\bm q}}$ is function which depends on full system states. $\hat{M}$ is the sysmetric, positive definite matrix. $\hat{ C}\left( \hat{\bm q}, \hat{\dot{\bm q}} \right)$ represents Christoffel symbols and $\hat{G}(\hat{\bm q})$ is the gravity vector. $\hat{\bm \lambda}$ represents ground reaction force. The generalized momentum of robot is 
\begin{equation*}
    \begin{aligned}
        \hat{\bm p} = M \, \hat{\dot{\bm q}}
    \end{aligned}
\end{equation*}
By differentiating the generalized momentum we get 
\begin{equation}
    \begin{aligned}
        \hat{\dot{ \bm p}} = \hat{M} \hat{\ddot{\bm q}} + \hat{\dot{M}}\hat{\dot{\bm q}}
    \end{aligned}
    \label{eq:change-momentum}
\end{equation}

Combining equation \eqref{eq:observer-dynamics} and \eqref{eq:change-momentum}
\begin{equation}
    \begin{aligned}
        \hat{\dot{ \bm p}} &= - \hat{\bm h} + \hat{\dot{M}}\hat{\dot{\bm q}} + \hat { B}_g \, \hat{\bm \lambda} + \hat{B}_t \, \hat{\bm u_t} +  \hat {B}_j \, \hat{\bm u_j}\\
        &= \hat{\bm \beta} + \bm r + \hat{B}_t \, \hat{\bm u_t} +  \hat {B}_j \, \hat{\bm u_j}
    \end{aligned}
\end{equation}

In this context, $\bm r = \hat{B}_t \, \hat{\bm u_t}$ represents the estimated thruster force. The term $\hat{\bm \beta}$ is defined as $\hat{\bm \beta} = -\hat{\bm h} + \hat{\dot{M}}\hat{\dot{\bm q}}$, where $\hat{\bm h}$ is derived from the Lagrangian dynamics, and  $\hat{\dot{M}}$ is numerically computed as $[\left(M_{k} - M_{k-1} \right)/\Delta t]$. Under ideal conditions, we assume $\hat{M} = M$ and $\hat{\bm \beta} = \bm \beta$. Consequently, the dynamic relationship between the estimated generalized thruster force $\bm r$ and the actual generalized thruster force $B_t \, \bm u_t$ is defined as


\begin{equation}
\begin{aligned}
     \dot{ \bm r} &= \bm K_0 \left( B_t \, \bm u_t - \bm r \right)\\
         &= \bm K_0 \left( \dot{\bm p}(t) - \hat{\dot{\bm p}}(t)\right)\\
\end{aligned}
\label{eq:estimator-dynamics}
\end{equation}
Equation \eqref{eq:estimator-dynamics} shows that $\bm r$ is low pass filter of thruster force. \cite{haddadin_robot_2017}, shows that as $ K_0 \rightarrow \infty \implies \bm r \approx B_t \, \bm u_t $. We can obtain the estimated GRF by integrating equation \eqref{eq:estimator-dynamics}
\begin{equation}
\begin{aligned}
    \bm r(t) &= \bm K_0 \left( \bm p(t) \int^{t}_{0} \left( \bm \beta + r(s) + \hat{B}_g \, \hat{\bm \lambda} +  \hat {B}_j \, \hat{\bm u_j}\right) ds - \hat{\bm p}(0)\right)
\end{aligned}
\label{eq:estimate-Gen-thruster-force}
\end{equation}
Further, it is possible to estimate the body frame thruster force by using jacobian  $J_t = [ J_{t,L}, J_{t,R}]$. 
\begin{equation}
\begin{aligned}
    \hat{\bm u}_t &= [J_{t}^{\top}J_t]^{\dagger}J_{t}^{\top} \bm r
\end{aligned}
\label{eq:estimate-thruster-force}
\end{equation}

From equation \eqref{eq:estimate-Gen-thruster-force}, it is evident that the estimation of thruster force necessitates the knowledge of ground reaction forces. There are two primary methodologies for estimating ground reaction forces: $(1)$ employing force sensors attached to the foot, or $(2)$ utilizing a contact constraint model. In the subsequent section, we will employ the contact constraint 

\subsection{Contact constraint model}
To estimate the ground reaction force, we constraint the stance foot acceleration to zero and we get $J_c\ddot{\bm q} = \dot{J}_c \dot{\bm q}$. By using this constraint and robot's dynamical acceleration \eqref{eq:eom_accel}, we get 

\begin{equation}
    \bm {\lambda} = \left( J_c M^{-1} J_c^{\top} \right)^{\dagger}\left( J_c M^{-1} (- \bm r - \hat {B}_j \, \hat{\bm u_j} + \bm h) - \dot{J}\dot{\bm q} \right) 
\label{eq:constraint-mdl}    
\end{equation}

where $(.)^{\dagger}$ is the Moore-Penrose pseudo-inverse and $J_c$ is the matrix of stacked foot contact Jacobians. During the double support phase, \(J_c M^{-1} J_c^{\top}\) is not full rank and this makes the estimation of the ground forces inherently inaccurate. 

\begin{figure}[t]
    \centering
    \includegraphics[width=1\linewidth]{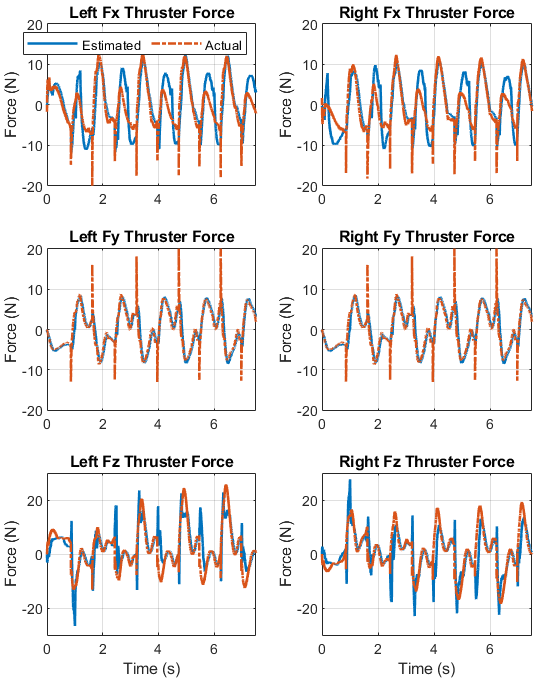}
    \caption{Illustrates body frame estimated thruster forces. GRF information is given by ground model}
    \label{fig:ut-ug}
\end{figure}
\begin{figure}[t]
    \centering
    \includegraphics[width=1\linewidth]{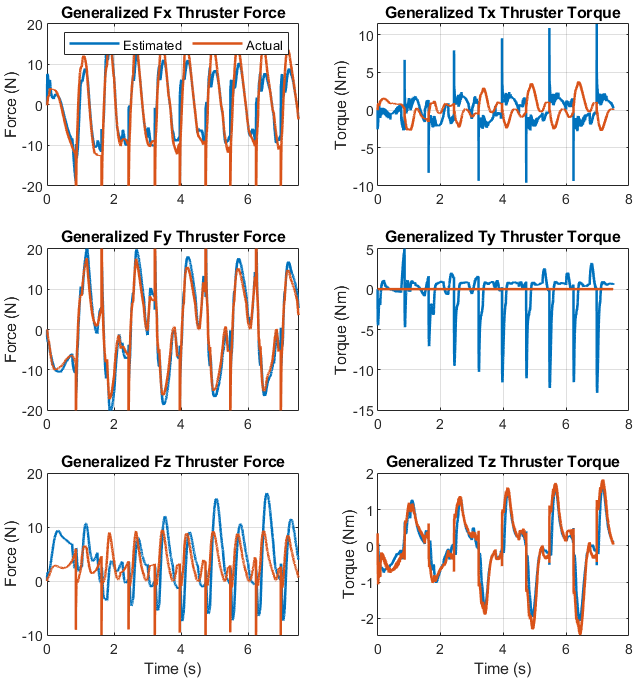}
    \caption{Illustrates estimated generalized thruster force from Conjugate momentum observer and terrain information from constraint model}
    \label{fig:Generalized-ut-lambda}
\end{figure}

\begin{figure}[t]
    \centering
    \includegraphics[width=1\linewidth]{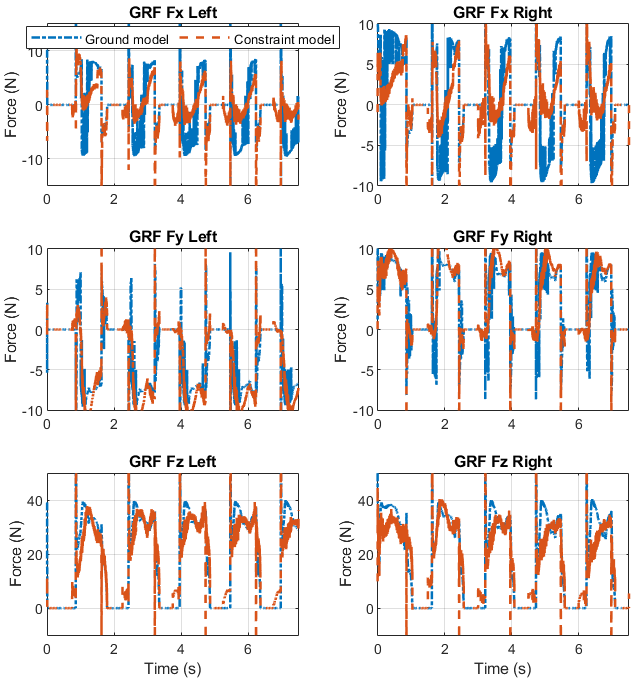}
    \caption{Comparison of GRF from ground model and Constraint model. Constraint model uses estimated generalized forces}
    \label{fig:GRF}
\end{figure}
\begin{figure}[t]
    \centering
    \includegraphics[width=1\linewidth]{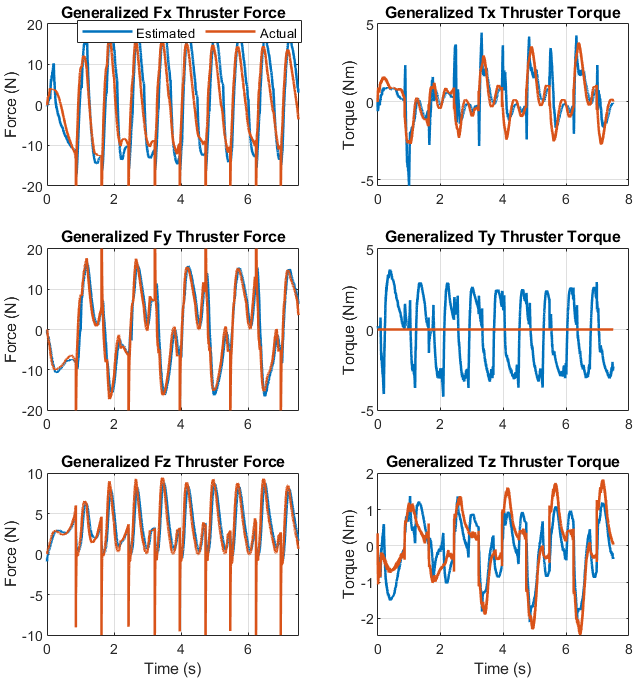}
    \caption{Illustrates estimated generalized thruster force from Conjugate momentum observer given terrain information from ground model}
    \label{fig:ut-ug}
\end{figure}
\begin{figure}[t]
    \centering
    \includegraphics[width=1\linewidth]{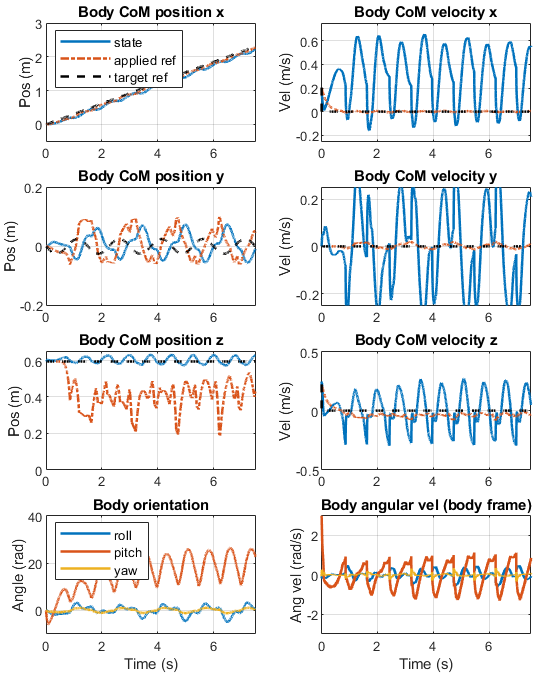}
    \caption{Illustrates the simulated robot states.}
    \label{fig:robot-state}
\end{figure}

\section{Results}
The numerical simulation and Observer design are done in the Matlab. We used RK4 scheme to propogate the model forward. Fig.~\ref{fig:robot-state} shows the position, velocity and orientation of the robot while performing thruster assisted walking.

\subsection{Simulation Specifications}
In this section, all units are in N, kg, m, s. The left leg of robot as following dimensions: $\bm l_1 = [0, 0.1, -0.1]^{\top}$, $\bm l_2 = [0, 0.5, 0]^{\top}$, $\bm l_3 = [0, 0, -0.3]^{\top}$ and $\bm l_4 = [0, 0.1, 0]^{\top}$. The right side has the $y$ axis component inverted. The following mass and inertias were used for simulation: $m_{B} = 2$, $m_{H} =m_{K} = 0.5$, $I_{B} = 10^{-3}$ and $I_{H} = I_{K} = 10^{-4}$. Finally, ground parameters were $\mu_s = 0.8$, $\mu_c = 0.64$, $\mu_v = 0.8$, $k_{g,p} = 8000$ and $k_{g,d} = 268$.

\subsection{Results and Discussion}
As discussed previously, To estimate the thruster force, we need to have information about the ground terrain.First method is to install force sensor on the foot which essentially allows to measure $\bm u_g$ directly. Fig.~\ref{fig:GRF} shows estimation of thruster force by inputing the ground reaction force from complaint ground model \eqref{eq:ground_model}. Observer gain used for estimation were $K_0 = diag(1,1,1,1,25,25,25,25,25,25)$. Generalized force and torque fairly matches apart from generalized torque in $y$ direction which is due to filter lag. Second method is to use constraint model. By using~\eqref{eq:estimate-thruster-force}, we found the body frame thruster force shown in fig.~\ref{fig:ut-ug}.

Fig.~\ref{fig:observer_flow_chart} show the flow chart for estimator dynamics when constraint model is used. The algorithm indicates that the constraint model relies on the estimated thruster force $\bm r$ and the observer dynamics terms $\hat{\beta}$. Consequently, any error in these estimations will result in a greater error in the ground reaction force estimation. To mitigate this, we can increase the observer gain. Fig.~\ref{fig:Generalized-ut-lambda} show the generalized thruster force estimation using constraint model~\eqref{eq:constraint-mdl} and Observer gain were set as $K_0 = diag(1,1,1,1,800,1200,60,3000,800,500)$. Generalized force in $F_x$, $F_y$ and $F_z$ matches fairly well, while in $\tau_x$ it oscillates about the actual. Generalized Torque $\tau_y$ and $\tau_z$ matches with closely with actual torque. Finally fig.~\ref{fig:GRF} shows the comparison between estimated GRF from constrained model with estimated thruster force $\bm r$ as input.

To evaluate the performance of our model, we calculated the Root Mean Square Error (RMSE) and subsequently normalized it to facilitate comparison across different datasets and scales. This normalization process yields a value between 0 and 1, where values closer to 0 indicate a better fit of the model to the data. By normalizing the RMSE, we ensure that the error metric is dimensionless and comparable across various contexts, providing a more robust assessment of model performance. Table \ref{tab:Norm-rmse} shows NRMSE for generalized forces and torque. This is for case where we use conjugate momentum with constraint model. The values are closer to zeros which shows the effectiveness of the method. The largest NRMSE is for $F_{z}$ which is due to use of constraint model.

\begin{table}[t]
    \centering
    \caption{Normalized RMSE value between actual and estimated generalized forces and torques.}
    \begin{tabular}{|c|c|}
        \hline
        Data & Normalized RMSE \\
        \hline
        $ Fx$ & 0.115586\\
        $ Fy$ & 0.049212\\
        $ Fz$ & 0.194566\\
        $ \tau_{x}$ & 0.110182\\
        $ \tau_{y}$ & 0.134201\\
        $ \tau_{z}$ & 0.123929\\
        \hline
    \end{tabular}
    \label{tab:Norm-rmse}
\end{table}

\section{conclusion}
In this work, we presented a conjugate momentum-based observer design for estimating thruster forces. Both the observer and simulation were running on a Matlab numerical simulator. We showed stable thruster-assisted walking simulation. While the robot is performing thruster-assisted walking, the observer is able to accurately tracked the thruster force. Since it is a low pass filter, the accuracy of the estimator highly depends on the ground reaction force information. The observer can accurately track the thruster forces given complete knowledge of ground forces. Alternative, we can also estimate the ground reaction force by using contact constraint. But, in doing so the accuracy of the estimator decreases. Our results demonstrated that the NRMSE values were consistently low, indicating a high degree of accuracy and reliability in our model’s predictions. This normalization process not only facilitated a more comprehensive evaluation but also highlighted the model’s effectiveness in various contexts
In future work, we will improve the accuracy of the estimator by introducing a second-order filter. Further, we will use these estimated thruster forces to perform dynamic maneuvers. 
%









\printbibliography

@inproceedings{mandralis_minimum_2023,
	title = {Minimum {Time} {Trajectory} {Generation} for {Bounding} {Flight}: {Combining} {Posture} {Control} and {Thrust} {Vectoring}},
	shorttitle = {Minimum {Time} {Trajectory} {Generation} for {Bounding} {Flight}},
	url = {https://ieeexplore.ieee.org/document/10178360},
	doi = {10.23919/ECC57647.2023.10178360},
	urldate = {2023-10-19},
	booktitle = {2023 {European} {Control} {Conference} ({ECC})},
	author = {Mandralis, Ioannis and Sihite, Eric and Ramezani, Alireza and Gharib, Morteza},
	month = jun,
	year = {2023},
	pages = {1--7},
}

@article{sihite_multi-modal_2023,
	title = {Multi-{Modal} {Mobility} {Morphobot} ({M4}) with appendage repurposing for locomotion plasticity enhancement},
	volume = {14},
	copyright = {2023 The Author(s)},
	issn = {2041-1723},
	url = {https://www.nature.com/articles/s41467-023-39018-y},
	doi = {10.1038/s41467-023-39018-y},
	language = {en},
	number = {1},
	urldate = {2023-07-04},
	journal = {Nature Communications},
	author = {Sihite, Eric and Kalantari, Arash and Nemovi, Reza and Ramezani, Alireza and Gharib, Morteza},
	month = jun,
	year = {2023},
	note = {Number: 1
Publisher: Nature Publishing Group},
	pages = {3323},
}

@inproceedings{liang_rough-terrain_2021,
	title = {Rough-{Terrain} {Locomotion} and {Unilateral} {Contact} {Force} {Regulations} {With} a {Multi}-{Modal} {Legged} {Robot}},
	doi = {10.23919/ACC50511.2021.9483189},
	booktitle = {2021 {American} {Control} {Conference} ({ACC})},
	author = {Liang, Kaier and Sihite, Eric and Dangol, Pravin and Lessieur, Andrew and Ramezani, Alireza},
	month = may,
	year = {2021},
	note = {ISSN: 2378-5861},
	pages = {1762--1769},
}

@inproceedings{ramezani_generative_2021,
	title = {Generative {Design} of {NU}’s {Husky} {Carbon}, {A} {Morpho}-{Functional}, {Legged} {Robot}},
	doi = {10.1109/ICRA48506.2021.9561196},
	booktitle = {2021 {IEEE} {International} {Conference} on {Robotics} and {Automation} ({ICRA})},
	author = {Ramezani, Alireza and Dangol, Pravin and Sihite, Eric and Lessieur, Andrew and Kelly, Peter},
	month = may,
	year = {2021},
	note = {ISSN: 2577-087X},
	pages = {4040--4046},
}

@inproceedings{sihite_optimization-free_2021,
	title = {Optimization-free {Ground} {Contact} {Force} {Constraint} {Satisfaction} in {Quadrupedal} {Locomotion}},
	doi = {10.1109/CDC45484.2021.9683155},
	booktitle = {2021 60th {IEEE} {Conference} on {Decision} and {Control} ({CDC})},
	author = {Sihite, Eric and Dangol, Pravin and Ramezani, Alireza},
	month = dec,
	year = {2021},
	note = {ISSN: 2576-2370},
	pages = {713--719},
}

@inproceedings{sihite_efficient_2022,
	title = {Efficient {Path} {Planning} and {Tracking} for {Multi}-{Modal} {Legged}-{Aerial} {Locomotion} {Using} {Integrated} {Probabilistic} {Road} {Maps} ({PRM}) and {Reference} {Governors} ({RG})},
	doi = {10.1109/CDC51059.2022.9992754},
	booktitle = {2022 {IEEE} 61st {Conference} on {Decision} and {Control} ({CDC})},
	author = {Sihite, Eric and Mottis, Benjamin and Ghanem, Paul and Ramezani, Alireza and Gharib, Morteza},
	month = dec,
	year = {2022},
	note = {ISSN: 2576-2370},
	pages = {764--770},
}

@article{dangol_control_2021,
	title = {Control of {Thruster}-{Assisted}, {Bipedal} {Legged} {Locomotion} of the {Harpy} {Robot}},
	volume = {8},
	issn = {2296-9144},
	url = {https://www.frontiersin.org/articles/10.3389/frobt.2021.770514},
	urldate = {2023-05-17},
	journal = {Frontiers in Robotics and AI},
	author = {Dangol, Pravin and Sihite, Eric and Ramezani, Alireza},
	year = {2021},
}

@inproceedings{dangol_feedback_2020-1,
	title = {Feedback design for {Harpy}: a test bed to inspect thruster-assisted legged locomotion},
	volume = {11425},
	shorttitle = {Feedback design for {Harpy}},
	url = {https://www.spiedigitallibrary.org/conference-proceedings-of-spie/11425/1142507/Feedback-design-for-Harpy--a-test-bed-to-inspect/10.1117/12.2558284.full},
	doi = {10.1117/12.2558284},
	urldate = {2023-05-17},
	booktitle = {Unmanned {Systems} {Technology} {XXII}},
	publisher = {SPIE},
	author = {Dangol, Pravin and Ramezani, Alireza},
	month = may,
	year = {2020},
	pages = {49--55},
}

@article{kim_bipedal_2021,
	title = {A bipedal walking robot that can fly, slackline, and skateboard},
	volume = {6},
	url = {https://www.science.org/doi/full/10.1126/scirobotics.abf8136},
	doi = {10.1126/scirobotics.abf8136},
	number = {59},
	urldate = {2022-10-30},
	journal = {Science Robotics},
	author = {Kim, Kyunam and Spieler, Patrick and Lupu, Elena-Sorina and Ramezani, Alireza and Chung, Soon-Jo},
	month = oct,
	year = {2021},
	note = {Publisher: American Association for the Advancement of Science},
	pages = {eabf8136},
}

@inproceedings{ruggiero_multilayer_2015,
	title = {A multilayer control for multirotor {UAVs} equipped with a servo robot arm},
	url = {https://ieeexplore.ieee.org/abstract/document/7139760},
	doi = {10.1109/ICRA.2015.7139760},
	urldate = {2024-10-02},
	booktitle = {2015 {IEEE} {International} {Conference} on {Robotics} and {Automation} ({ICRA})},
	author = {Ruggiero, F. and Trujillo, M.A. and Cano, R. and Ascorbe, H. and Viguria, A. and Peréz, C. and Lippiello, V. and Ollero, A. and Siciliano, B.},
	month = may,
	year = {2015},
	note = {ISSN: 1050-4729},
	pages = {4014--4020},
}

@article{mohamed_momentum-based_2022,
	title = {Momentum-{Based} {Extended} {Kalman} {Filter} for {Thrust} {Estimation} on {Flying} {Multibody} {Robots}},
	volume = {7},
	issn = {2377-3766},
	url = {https://ieeexplore.ieee.org/document/9622189/?arnumber=9622189},
	doi = {10.1109/LRA.2021.3129258},
	number = {1},
	urldate = {2024-10-02},
	journal = {IEEE Robotics and Automation Letters},
	author = {Mohamed, Hosameldin Awadalla Omer and Nava, Gabriele and L’Erario, Giuseppe and Traversaro, Silvio and Bergonti, Fabio and Fiorio, Luca and Vanteddu, Punith Reddy and Braghin, Francesco and Pucci, Daniele},
	month = jan,
	year = {2022},
	note = {Conference Name: IEEE Robotics and Automation Letters},
	pages = {526--533},
}

@inproceedings{cai_predefined-time_2023,
	address = {Singapore},
	title = {Predefined-{Time} {External} {Force} {Estimation} for {Legged} {Robots}},
	isbn = {978-981-9964-95-6},
	doi = {10.1007/978-981-99-6495-6_46},
	language = {en},
	booktitle = {Intelligent {Robotics} and {Applications}},
	publisher = {Springer Nature},
	author = {Cai, Peiyuan and Liu, Danfu and Zhu, Lijun},
	editor = {Yang, Huayong and Liu, Honghai and Zou, Jun and Yin, Zhouping and Liu, Lianqing and Yang, Geng and Ouyang, Xiaoping and Wang, Zhiyong},
	year = {2023},
	pages = {542--552},
}

@inproceedings{pitroda_capture_2024,
	title = {Capture {Point} {Control} in {Thruster}-{Assisted} {Bipedal} {Locomotion}},
	url = {https://ieeexplore.ieee.org/abstract/document/10637139?casa_token=caq4axqZ51QAAAAA:hBX7LRUriIsppyQYfbQzjCm1sipqibztR-RjvG-uWgN4nqs-R1zzU7EMlnKHr5uUKfEtcJ0W},
	doi = {10.1109/AIM55361.2024.10637139},
	urldate = {2024-09-29},
	booktitle = {2024 {IEEE} {International} {Conference} on {Advanced} {Intelligent} {Mechatronics} ({AIM})},
	author = {Pitroda, Shreyansh and Bondada, Aditya and Venkatesh, Kaushik and Salagame, Adarsh and Wang, Chenghao and Liu, Taoran and Gupta, Bibek and Sihite, Eric and Nemovi, Reza and Ramezani, Alireza and Gharib, Morteza},
	month = jul,
	year = {2024},
	note = {ISSN: 2159-6255},
	pages = {1139--1144},
}

@misc{noauthor_making_nodate,
	title = {On making robots understand safety: {Embedding} injury knowledge into control - {Sami} {Haddadin}, {Simon} {Haddadin}, {Augusto} {Khoury}, {Tim} {Rokahr}, {Sven} {Parusel}, {Rainer} {Burgkart}, {Antonio} {Bicchi}, {Alin} {Albu}-{Schäffer}, 2012},
	url = {https://journals.sagepub.com/doi/abs/10.1177/0278364912462256?casa_token=zT9wDmGIPrMAAAAA:gr3D3cWNbt9AaTxY3OviIWYYZ94EguNmL_z5KOrOvcj08bazl1xbWYyj9PMjj9wGENT6nr71fLY},
	urldate = {2024-09-29},
}

@inproceedings{garofalo_sliding_2019,
	title = {Sliding {Mode} {Momentum} {Observers} for {Estimation} of {External} {Torques} and {Joint} {Acceleration}},
	url = {https://ieeexplore.ieee.org/document/8793529/?arnumber=8793529},
	doi = {10.1109/ICRA.2019.8793529},
	urldate = {2024-09-29},
	booktitle = {2019 {International} {Conference} on {Robotics} and {Automation} ({ICRA})},
	author = {Garofalo, Gianluca and Mansfeld, Nico and Jankowski, Julius and Ott, Christian},
	month = may,
	year = {2019},
	note = {ISSN: 2577-087X},
	pages = {6117--6123},
}

@article{tomic_external_2017,
	title = {External {Wrench} {Estimation}, {Collision} {Detection}, and {Reflex} {Reaction} for {Flying} {Robots}},
	volume = {33},
	issn = {1941-0468},
	url = {https://ieeexplore.ieee.org/document/8059847/?arnumber=8059847},
	doi = {10.1109/TRO.2017.2750703},
	number = {6},
	urldate = {2024-09-29},
	journal = {IEEE Transactions on Robotics},
	author = {Tomić, Teodor and Ott, Christian and Haddadin, Sami},
	month = dec,
	year = {2017},
	note = {Conference Name: IEEE Transactions on Robotics},
	pages = {1467--1482},
}

@inproceedings{de_luca_actuator_2003,
	title = {Actuator failure detection and isolation using generalized momenta},
	volume = {1},
	url = {https://ieeexplore.ieee.org/document/1241665/?arnumber=1241665},
	doi = {10.1109/ROBOT.2003.1241665},
	urldate = {2024-09-29},
	booktitle = {2003 {IEEE} {International} {Conference} on {Robotics} and {Automation} ({Cat}. {No}.{03CH37422})},
	author = {De Luca, A. and Mattone, R.},
	month = sep,
	year = {2003},
	note = {ISSN: 1050-4729},
	pages = {634--639 vol.1},
}

@article{liu_sensorless_2024,
	title = {Sensorless {Ground} {Reaction} {Force} {Observation} {With} {Disturbance} {Compensation} in {Heavy}-{Legged} {Robots}},
	issn = {1941-014X},
	url = {https://ieeexplore.ieee.org/document/10423296/?arnumber=10423296},
	doi = {10.1109/TMECH.2024.3354989},
	urldate = {2024-09-29},
	journal = {IEEE/ASME Transactions on Mechatronics},
	author = {Liu, Shaoxun and Pan, Zheng and Zhou, Shiyu and Niu, Zhihua and Wang, Rongrong},
	year = {2024},
	note = {Conference Name: IEEE/ASME Transactions on Mechatronics},
	pages = {1--12},
}

@misc{noauthor_residual-based_nodate,
	title = {Residual-based contacts estimation for humanoid robots {\textbar} {IEEE} {Conference} {Publication} {\textbar} {IEEE} {Xplore}},
	url = {https://ieeexplore.ieee.org/abstract/document/7803308?casa_token=2RhORD0rn4IAAAAA:3fe9evpFqIEVpAHm23ly55x8MvR9ff_HwCPHTNpoT10zm5e7xUHYW38kAIiSKG5qSoe9kIbr},
	urldate = {2024-09-29},
}

@inproceedings{de_luca_sensorless_2005,
	title = {Sensorless {Robot} {Collision} {Detection} and {Hybrid} {Force}/{Motion} {Control}},
	url = {https://ieeexplore.ieee.org/abstract/document/1570247},
	doi = {10.1109/ROBOT.2005.1570247},
	urldate = {2024-09-28},
	booktitle = {Proceedings of the 2005 {IEEE} {International} {Conference} on {Robotics} and {Automation}},
	author = {de Luca, A. and Mattone, R.},
	month = apr,
	year = {2005},
	note = {ISSN: 1050-4729},
	pages = {999--1004},
}

@inproceedings{de_luca_collision_2006,
	title = {Collision {Detection} and {Safe} {Reaction} with the {DLR}-{III} {Lightweight} {Manipulator} {Arm}},
	url = {https://ieeexplore.ieee.org/document/4058607/?arnumber=4058607},
	doi = {10.1109/IROS.2006.282053},
	urldate = {2024-08-08},
	booktitle = {2006 {IEEE}/{RSJ} {International} {Conference} on {Intelligent} {Robots} and {Systems}},
	author = {De Luca, Alessandro and Albu-Schaffer, Alin and Haddadin, Sami and Hirzinger, Gerd},
	month = oct,
	year = {2006},
	note = {ISSN: 2153-0866},
	pages = {1623--1630},
}

@inproceedings{bledt_contact_2018,
	title = {Contact {Model} {Fusion} for {Event}-{Based} {Locomotion} in {Unstructured} {Terrains}},
	url = {https://ieeexplore.ieee.org/document/8460904/?arnumber=8460904},
	doi = {10.1109/ICRA.2018.8460904},
	urldate = {2024-08-08},
	booktitle = {2018 {IEEE} {International} {Conference} on {Robotics} and {Automation} ({ICRA})},
	author = {Bledt, Gerardo and Wensing, Patrick M. and Ingersoll, Sam and Kim, Sangbae},
	month = may,
	year = {2018},
	note = {ISSN: 2577-087X},
	pages = {4399--4406},
}

@article{haddadin_robot_2017,
	title = {Robot {Collisions}: {A} {Survey} on {Detection}, {Isolation}, and {Identification}},
	volume = {33},
	issn = {1941-0468},
	shorttitle = {Robot {Collisions}},
	url = {https://ieeexplore.ieee.org/document/8059840/?arnumber=8059840&tag=1},
	doi = {10.1109/TRO.2017.2723903},
	number = {6},
	urldate = {2024-07-18},
	journal = {IEEE Transactions on Robotics},
	author = {Haddadin, Sami and De Luca, Alessandro and Albu-Schäffer, Alin},
	month = dec,
	year = {2017},
	note = {Conference Name: IEEE Transactions on Robotics},
	pages = {1292--1312},
}

@article{pitroda_dynamic_2023,
	title = {Dynamic multimodal locomotion: a quick overview of hardware and control.},
	shorttitle = {Dynamic multimodal locomotion},
	url = {https://repository.library.northeastern.edu/files/neu:4f21z897z},
	urldate = {2024-07-08},
	author = {Pitroda, Shreyansh},
	year = {2023},
}

@article{dangol_control_2021-1,
	title = {Control of {Thruster}-{Assisted}, {Bipedal} {Legged} {Locomotion} of the {Harpy} {Robot}},
	volume = {8},
	doi = {10.3389/frobt.2021.770514},
	journal = {Frontiers in Robotics and AI},
	author = {Dangol, Pravin and Sihite, Eric and Ramezani, Alireza},
	month = dec,
	year = {2021},
}

\end{document}